\documentclass{article}

\usepackage{PRIMEarxiv}

\usepackage[utf8]{inputenc} 
\usepackage[T1]{fontenc}    
\usepackage{hyperref}       
\usepackage{url}            
\usepackage{booktabs}       
\usepackage{amsfonts}       
\usepackage{nicefrac}       
\usepackage{microtype}      
\usepackage{lipsum}
\usepackage{fancyhdr}       
\usepackage{graphicx}       
\graphicspath{{media/}}     

\usepackage{natbib}
\usepackage{adjustbox}
\usepackage{bbm}
\usepackage{bm}
\usepackage{amsmath}
\usepackage{booktabs}

\usepackage{makecell}
\usepackage{caption}
\usepackage{subcaption}
\usepackage{multirow}
\usepackage{xurl}
\usepackage[dvipsnames]{xcolor}

\title{GreenKGC: A Lightweight Knowledge Graph Completion Method}

\author{
  Yun-Cheng Wang \\
  University of Southern California \\
  Los Angeles, USA\\
  \texttt{yunchenw@usc.edu} \\
   \And
  Xiou Ge \\
  University of Southern California \\
  Los Angeles, USA\\
  \texttt{xiouge@usc.edu} \\
   \And
  Bin Wang \\
  National University of Singapore \\
  Singapore\\
  \texttt{bwang28c@gmail.com} \\
   \And
  C.-C. Jay Kuo \\
  University of Southern California \\
  Los Angeles, USA\\
  \texttt{cckuo@sipi.usc.edu} \\
}

\begin{document}
\maketitle

\begin{abstract}
Knowledge graph completion (KGC) aims to discover missing relationships
between entities in knowledge graphs (KGs).  Most prior KGC work focuses
on learning embeddings for entities and relations through a simple 
scoring function. Yet, a higher-dimensional embedding space 
is usually required for a better reasoning capability, which 
leads to a larger model size and hinders applicability to 
real-world problems (e.g., large-scale KGs or mobile/edge computing). 
A lightweight modularized KGC solution, called GreenKGC, is 
proposed in this work to address this issue.  GreenKGC
consists of three modules: representation learning, feature
pruning, and decision learning, to extract discriminant KG features
and make accurate predictions on missing relationships using classifiers
and negative sampling.
Experimental results demonstrate that, in low dimensions, GreenKGC can
outperform SOTA methods in most datasets. In addition, low-dimensional 
GreenKGC can achieve competitive or even better performance against
high-dimensional models with a much smaller model size. We make our
code publicly available.\footnote{\url{https://github.com/yunchengwang/GreenKGC}}
\end{abstract}


\section{Introduction}\label{sec:introduction}

Knowledge graphs (KGs) store human knowledge in a graph-structured
format, where nodes and edges denote entities and relations,
respectively. A (\emph{head entity}, \emph{relation}, \emph{tail
entity}) factual triple, denoted by $(h, r, t)$, is a basic
component in KGs. In many knowledge-centric artificial intelligence (AI)
applications, such as question answering \citep{huang2019knowledge,
saxena2020improving}, information extraction 
\citep{hoffmann2011knowledge, daiber2013improving}, 
and recommendation \citep{wang2019explainable,
xian2019reinforcement}, KG plays an important role as it provides
explainable reasoning paths to predictions. However, most KGs suffer
from the incompleteness problem; namely, a large number of factual
triples are missing, leading to performance degradation in downstream
applications. Thus, there is growing interest in developing KG
completion (KGC) methods to solve the incompleteness problem by inferring 
undiscovered factual triples based on existing ones. 
\begin{figure}[t]
\centering
\includegraphics[width=0.6\textwidth]{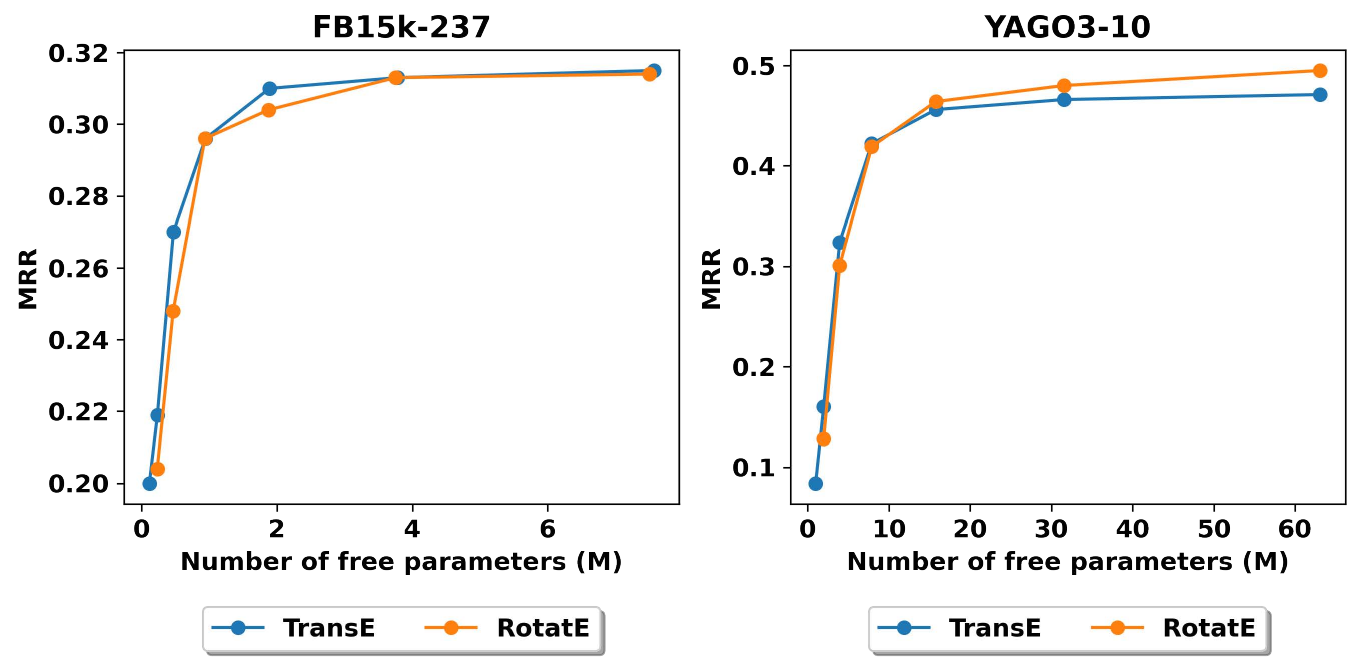}
\caption{MRR versus the number of free parameters in
KGE methods against FB15K-237 (left) and YAGO3-10
dataset (right). When a model has fewer parameters, its performance is
poorer. Also, the larger dataset, YAGO3-10, demands more
parameters than the smaller dataset, FB15k-237, to achieve satisfactory results.}
\label{fig:dimension}
\end{figure}
Knowledge graph embedding (KGE) methods have been widely used to solve
the incompleteness problem. Embeddings for entities and relations are
stored as model parameters and updated by maximizing triple scores among
observed triples while minimizing those among negative triples. The number of
free parameters in a KGE model is linear to the embedding dimension and the
number of entities and relations in KGs, i.e. $O((|E| + |R|)d)$, where
$|E|$ is the number of entities, $|R|$ is the number of relations, and
$d$ is the embedding dimension. Since KGE models usually require a
higher-dimensional embedding space for a better reasoning capability,
they require large model sizes (i.e. parameter numbers) to achieve
satisfactory performance as demonstrated in Fig.~\ref{fig:dimension}.
To this end, it is challenging for them to handle large-scale KGs with
lots of entities and relations in resource-constrained platforms such as
mobile/edge computing. A KGC method that has good reasoning capability
in low dimensions is desired \cite{kuo2022green}. 

The requirement of high-dimensional embeddings for popular KGE methods comes
from the over-simplified scoring functions \citep{xiao2015transa}.
Thus, classification-based KGC methods, such as ConvE
\citep{dettmers2018convolutional}, aim to increase the reasoning
capabilities in low dimensions by adopting neural networks (NNs)
as powerful decoders. As a result, they are more efficient in parameter 
scaling than KGE models \citep{dettmers2018convolutional}. However, NNs 
demand longer inference time and more computation power due to their deep 
architectures. The long inference time of the classification-based methods also 
limits their applicability to some tasks that require real-time inference. 
Recently, DualDE \citep{zhu2022dualde} applied Knowledge Distillation (KD) 
\citep{hinton2015distill} to train powerful low-dimensional embeddings. 
Yet, it demands three stages of embedding training: 1) training
high-dimensional KGE, 2) training low-dimensional KGE with the guidance
of high-dimensional KGE, and 3) multiple rounds of student-teacher
interactions. Its training process is time-consuming and may fail to 
converge when the embeddings are not well-initialized. 

Here, we propose a new KGC method that works well under low dimensions
and name it GreenKGC. GreenKGC consists of three modules: 1)
representation learning, 2) feature pruning, and 3) decision learning.
Each of them is trained independently. In Module 1, we leverage a KGE
method, called the baseline method, to learn high-dimensional entity and
relation representations. In Module 2, a feature pruning process is
applied to the high-dimensional entity and relation representations to
yield discriminant low-dimensional features for triples. In addition, we
observe that some feature dimensions are more powerful than others in 
different relations. Thus, we group relations with similar discriminant 
feature dimensions for parameter savings and better performance.
In Module 3, we train a binary classifier for each relation group so that it can
predict triple's score in inference. The score is a soft prediction
between 0 and 1, which indicates the probability of whether a certain
triple exists or not. Finally, we propose two novel negative sampling schemes, 
embedding-based and ontology-based, for classifier training in this work.
They are used for hard negative mining, where these hard negatives cannot be 
correctly predicted by the baseline KGE methods. 

We conduct extensive
experiments and compare the performance and model sizes of GreenKGC with
several representative KGC methods on link prediction datasets. 
Experimental results show that GreenKGC can achieve good performance
in low dimensions, i.e. 8, 16, 32 dimensions, compared with SOTA 
low-dimensional methods. In addition, GreenKGC shows competitive or
better performance compared to the high-dimensional KGE methods 
with a much smaller model size. We also conduct experiments on
a large-scale link prediction datasets with over 2.5M entities
and show that GreenKGC can perform well with much fewer model 
parameters.
Ablation studies are also conducted to show 
the effectiveness of each module in GreenKGC.

\renewcommand{\figurename}{Figure}
\renewcommand{\tablename}{Table}

\section{Related Work}\label{sec:review}

\subsection{KGE Methods}

Distance-based KGE methods model relations as affine transformations from head entities to
tail entities. For example, TransE \citep{bordes2013translating} models relations as 
translations, while RotatE \citep{sun2018rotate} models relations as rotations in 
the complex embedding space for better expressiveness on symmetric relations. 
Recent work has tried to model relations as scaling \citep{chao2021pairre} and reflection 
\citep{zhang2022knowledge} operations in order to handle particular relation patterns. 
Semantic-matching KGE methods, such as RESCAL \citep{lin2015learning} and DistMult 
\citep{bordes2014semantic}, formulate the scoring functions as similarities 
among head, relation, and tail embeddings. ComplEx~\citep{trouillon2016complex} 
extends such methods to a complex space for better expressiveness on asymmetric relations. 
Recently, TuckER~\citep{balazevic2019tucker} and AutoSF~\citep{zhang2020autosf} allow 
more flexibility in modeling similarities. Though KGE methods are simple, 
they often require a high-dimensional embedding space to be expressive.

\subsection{Classification-based KGC Methods}

NTN \citep{socher2013reasoning} adopts a neural 
tensor network combined with textual representations of entities. 
ConvKB \citep{nguyen2018novel} uses $1 \times 3$
convolutional filters followed by several fully connected
(FC) layers to predict triple scores.  ConvE~\citep{dettmers2018convolutional} reshapes entity and relation
embeddings into 2D images and uses $3 \times 3$ convolutional filters
followed by several FC layers to predict the scores of triples. 
Though NN-based methods can achieve good performance in a lower dimension,
they have several drawbacks, such as long inference time and large model. 
KGBoost \citep{wang2022kgboost} is a classification-based method that doesn't
use NNs. Yet, it assigns one classifier for each relation so it's not scalable
to large-scale datasets.

\subsection{Low-dimensional KGE Methods}

Recently, research on the design of low-dimensional KGE methods has
received attention. MuRP \citep{balazevic2019multi} embeds entities
and relations in a hyperbolic space due to its effectiveness in modeling 
hierarchies in KGs. AttH \citep{chami2020low} improves hyperbolic KGE 
by leveraging hyperbolic isometries to model logical patterns. MulDE 
\citep{wang2021mulde} adopts Knowledge Distillation \citep{hinton2015distill} 
on a set of hyperbolic KGE as teachers to learn powerful embeddings in low dimensions.
However, embeddings in hyperbolic space are hard to be used in other downstream tasks.
In Euclidean space, DualDE~\citep{zhu2022dualde} adopts Knowledge Distillation 
to learn low-dimensional embeddings from high-dimensional ones for smaller model
sizes and faster inference time. Yet, it requires a long training time
to reduce feature dimension.
GreenKGC has two clear advantages over existing low-dimensional methods. 
First, it fully operates in the Euclidean space. Second, it does not 
need to train new low-dimensional embeddings from scratch, thus requiring
a shorter dimension reduction time. 

\begin{figure*}[t]
\centering
\includegraphics[width=0.96\textwidth]{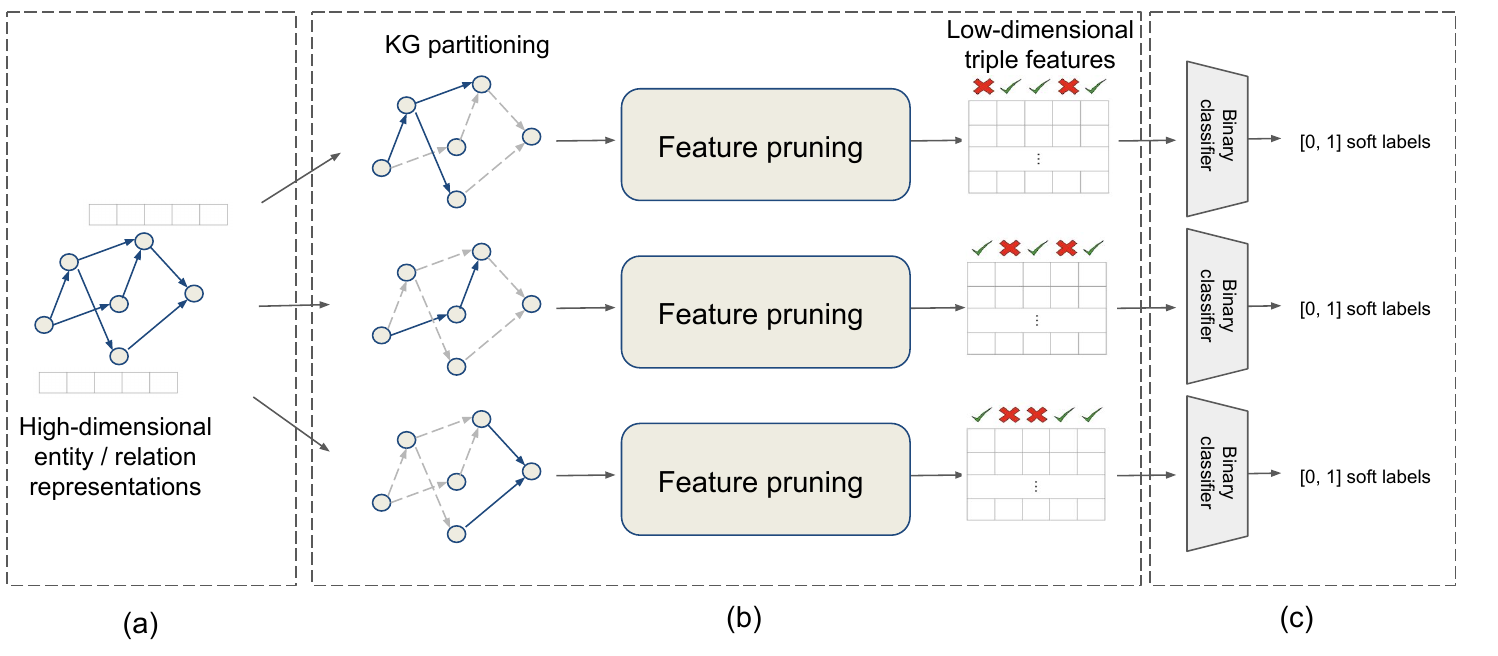}
\caption{An overview of GreenKGC, which consists of three modules: (a)
representation learning, (b) feature pruning, and (c) decision learning.} 
\label{fig:overview}
\end{figure*}

\section{Methodology}\label{sec:method}

GreenKGC is presented in this section. It consists of three modules:
representation learning, feature pruning, and decision learning, to
obtain discriminant low-dimensional triple features and predict triple
scores accurately.  An overview of GreenKGC is given in Fig.~\ref{fig:overview}. Details of each module will be elaborated below. 

\subsection{Representation Learning}\label{subsec:representation}

We leverage existing KGE models, such as TransE~\citep{bordes2013translating} 
and RotatE~\citep{sun2018rotate}, to obtain good initial embeddings for entities
and relations, where their embedding dimensions can be high to be
expressive. Yet, the initial embedding dimension will be largely reduced
in the feature pruning module. In general, GreenKGC can build upon any
existing KGE models. We refer to the KGE models used in GreenKGC as our 
baseline models.
We include the training details for baseline models 
in Appendix \ref{appendix:emb} as they are not the main focus of this paper.


\subsection{Feature Pruning}\label{subsec:feature}

In this module, a small subset of feature dimensions in high-dimensional
KG representations from Module 1 are preserved, while the others are pruned,
to form low-dimensional discriminant KG features.

\textbf{Discriminant Feature Test (DFT).} DFT is a supervised feature
selection method recently proposed in \citet{yang2022supervised}.  All
training samples have a high-dimensional feature set as well as the
corresponding labels. DFT scans through each dimension in the feature
set and computes its discriminability based on sample labels.  DFT can
be used to reduce the dimensions of entity and relation embeddings while
preserving their power in downstream tasks such as KGC. 

Here, we extend DFT to the multivariate setting since there are
multiple variables in each triple. For example, TransE 
\citep{bordes2013translating} has 3 variables (i.e. $\bm{h}$, $\bm{r}$, and $\bm{t}$) in 
each feature dimension. 
First, for each dimension $i$, we learn a linear transformation $\bm{w}_i$ to map 
multiple variables $[h_i, r_i, t_i]$ to a 
single variable $x_i$ in each triple, where $h_i, r_i, t_i$ represents the 
$i$-th dimension in the head, relation, and tail representations, respectively. Such a linear 
transformation can be 
learned through principal component analysis (PCA) using singular value 
decomposition (SVD). As a result, $\bm{w}_i$ is the first principal component in PCA.
However, linear transformations learned from PCA are unsupervised and cannot 
separate observed triples from negatives well. Alternatively, we learn the linear 
transformation through logistic regression by minimizing the binary cross-entropy loss

\begin{equation}\label{equ:logistic}
\begin{split}
        \mathcal{L} = {} & -y \log(\sigma(\bm{w}_i[h_i, r_i, t_i]^T)) \\
                         & -(1 - y)
                    \log(1 - \sigma(\bm{w}_i[h_i, r_i, t_i]^T)), 
\end{split}
\end{equation}

where $y = 1$ for observed triples $(h, r, t)$ and $y = 0$ for corrupted triples $(h', r, t')$. 
Afterward, we can apply the standard DFT to each dimension.

DFT adopts cross-entropy (CE) to evaluate the discriminant power of each dimension as
CE is a typical loss for binary classification.  Dimensions
with lower CE imply higher discriminant power. We preserve the 
feature dimensions with the lowest CE and prune the remaining 
to obtain low-dimensional features. Details for training DFT
are given in Appendix \ref{appendix:dft}.


\textbf{KG partitioning.} Given that relations in KGs could be different 
(e.g. symmetric v.s. asymmetric and \emph{films} v.s. \emph{sports}), a small subset of feature dimensions
might not be discriminant for all relations. Thus, we first partition 
them into disjoint relations groups, where relations in each group 
have similar properties. Then, we perform feature pruning within each 
relation group and select the powerful feature dimensions correspondingly.

\begin{table}[t]
\centering
\begin{tabular}{ c l }
\toprule
Cluster \# & Relations \\
\midrule
0 & \_derivationally\_related\_form \\
  & \_also\_see \\
  & \_member\_meronym \\
  & \_has\_part \\
  & \_verb\_group \\
  & \_similar\_to \\
\midrule
1 & \_hypernym \\
  & \_instance\_hypernym \\
  & \_synset\_domain\_topic\_of \\
\midrule
2 & \_member\_of\_domain\_usage \\
  & \_member\_of\_domain\_region \\
\bottomrule
\end{tabular}
\caption{Relation grouping results on WN18RR when applying $k$-Means on 
relation embeddings when $k$ = 3.}
\label{tab:partition}
\end{table}

We hypothesize that relations that have similar properties are close in the 
embedding space. Therefore, we use $k$-Means to cluster relation embeddings
into relation groups. To verify our hypothesize, we show the grouping results on
WN18RR in Table \ref{tab:partition}. Without categorizing relations into 
different logical patterns explicitly, relations of similar patterns can be 
clustered together in the embedding space. For example, most
relations in cluster \#0 are symmetric ones. All relations in the cluster
\#1 are N-to-1. The remaining two relations in cluster \#2 are 1-to-N
with the highest tail-per-head ratio. While we observe cardinality-based
grouping for relations in WN18RR, which mostly contains abstract concepts, 
for FB15k-237 and YAGO3-10, relations with similar semantic 
meanings are often grouped after KG partitioning.

\begin{figure}[t]
\centering
\includegraphics[width=0.6\textwidth]{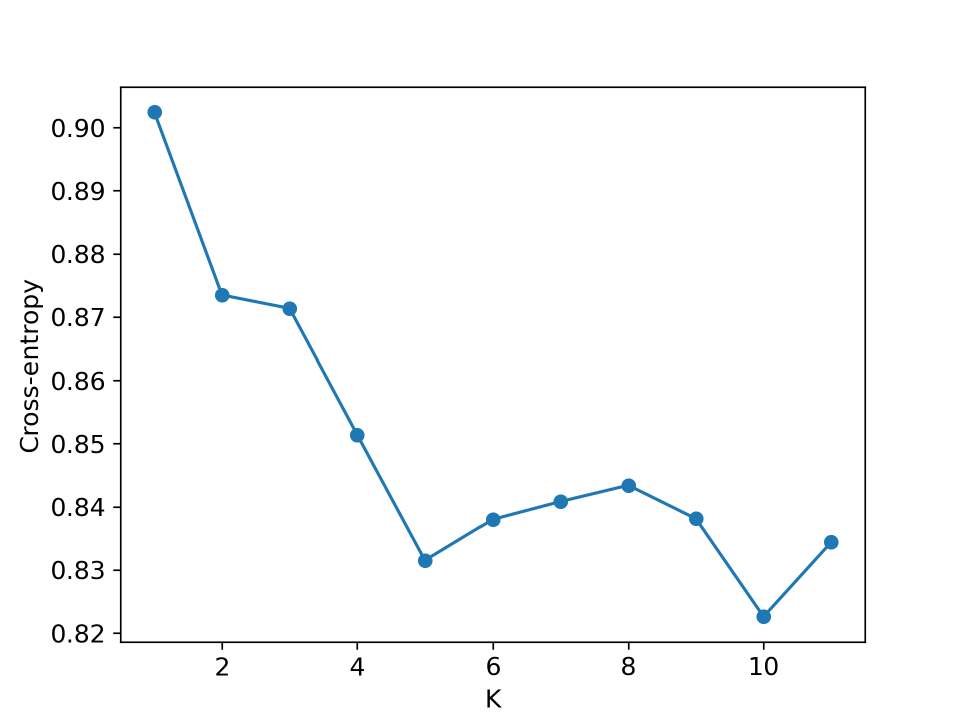} \\
\caption{Average cross-entropy for different numbers of KG partitions
in FB15k-237.}\label{fig:avg_ce}
\end{figure}

Furthermore, we evaluate how different numbers of relation groups, $k$, can affect the feature 
pruning process. In Fig. \ref{fig:avg_ce}, as the lower CE reflects more 
discriminant features, we can obtain more powerful features when $k$ becomes larger, i.e. 
partitioning KG into more relation groups. 
Thus, for each dataset, we select the optimal $k$ when the average CE starts to converge. We 
elaborate on the high-level intuition on why combining feature pruning and KG partitioning 
works with KGE models. First, KGE models are isotropic, meaning each dimension can be handled 
by DFT independently. Second, some feature dimensions are more powerful than others in 
different relations. Thus, we group relations that with the same discriminant feature 
dimensions for parameter savings.


\subsection{Decision Learning}\label{subsec:decision}

We formulate KGC as a binary classification problem in each
relation group. We adopt binary classifiers as decoders since they are
more powerful than simple scoring functions. The binary
classifiers take pruned triple features as inputs and predict
soft probabilities (between 0 and 1) of triples as outputs. We also
conduct classifier training with hard negative mining so as to train a
powerful classifier. 

\textbf{Binary classification.} The binary classifiers, $g(*)$, take a
low-dimensional triple feature $\bm{x}$ and predict a soft label
$\hat{y} = g(\bm{x}) \in [0, 1]$. The label $y = 1$ for the observed triples and $y = 0$
for the sampled negatives. We train a binary classifier by minimizing the following
negative log-likelihood loss:
\begin{equation}
\begin{split}
        l(y, \hat{y}) = {} & - y \log(\hat{y}) \\
                           & - (1 - y) \log(1 - \hat{y}), 
\end{split}
\end{equation}
In general, we select a nonlinear classifier to accommodate 
nonlinearity in sample distributions. 

\textbf{Negative sampling.} Combining KGE with classifiers is non-trivial 
because it's challenging to obtain high-quality negative samples for classifier
training, given that negative samples are not explicitly labeled in the KGs. 
Therefore, it is desired to mine hard negative cases for baseline KGE models
so as to train a powerful classifier. 
We propose two negative sampling schemes for classifier training.
First, most KGE models can only capture the
coarse entity type information. For example, they may predict a location
given the query (\emph{Mary}, \emph{born\_in}, \emph{?}) yet
without an exact answer. Thus, we draw negative samples within the
entity types constrained by relations \citep{krompass2015type}
to enhance the capability to predict the exact answer.  
Such a negative sampling scheme is called
\emph{ontology-based negative sampling}.  We also investigate the sampling of
hard negatives that cannot be trivially obtained from original KGE
methods. Negatives with higher embedding scores $f_r(\bm{h_i},
\bm{t_i})$ tend to be predicted wrongly in the baseline methods. To handle
it, we rank all randomly sampled negative triples and select the ones
with higher embedding scores as hard negatives for classifier training.
Such a negative sampling strategy is called \emph{embedding-based negative sampling}.

\section{Experiments}\label{sec:experiment}

\begin{table}[!t]
\color{black}
\setlength\tabcolsep{3pt}
\centering
\begin{tabular}{ c c c c }
\toprule
Dataset & \# ent. & \# rel. & \# triples (train / valid / test) 
\\
\midrule
WN18RR    &40,943 &11 &86,835 / 3,034 / 3,134 
\\
FB15k-237 &14,541 &237 &272,115 / 17,535 / 20,466 
\\
YAGO3-10 &123,143 &37 &1,079,040 / 4,978 / 4,982
\\ \midrule
ogbl-wikikg2 & 2,500,604 & 535 & 16,109,182 / 429,456 / 598,543
\\
\bottomrule
\end{tabular}
\caption{Dataset statistics.}
\label{table:dataset}
\end{table}


\begin{table*}[!t]
\setlength\tabcolsep{3pt}
\centering
\resizebox*{\textwidth}{!}{
\begin{tabular}{l | c  c  c  c | c  c  c  c | c  c  c  c}
\hline
& \multicolumn{4}{c|}{\textbf{FB15k-237}} & \multicolumn{4}{c|}{\textbf{WN18RR}} & \multicolumn{4}{c}{\textbf{YAGO3-10}} \\
\hline
Model & MRR & H@1 & H@3 & H@10 & MRR & H@1 & H@3 & H@10 & MRR & H@1 & H@3 & H@10 \\
\hline
\multicolumn{13}{l}{\emph{KGE Methods}} \\
\hline
TransE \citep{bordes2013translating} & 0.270 & 0.177 & 0.303 & 0.457 
      & 0.150 & 0.009 & 0.251 & 0.387
      & 0.324 & 0.221 & 0.374 & 0.524 \\
RotatE \citep{sun2018rotate} & 0.290 & 0.208 & 0.316 & 0.458 
       & 0.387 & 0.330 & 0.417 & 0.491 
       & \underline{0.419} & \underline{0.321} & \underline{0.475} & \underline{0.607} \\
\hline
\multicolumn{13}{l}{\emph{Classification-based Methods}} \\
\hline
ConvKB \citep{nguyen2018novel} & 0.232 & 0.157 & 0.255 & 0.377
       & 0.346 & 0.300 & 0.374 & 0.422 
       & 0.311 & 0.194 & 0.368 & 0.526 \\
ConvE  \citep{dettmers2018convolutional} & 0.282 & 0.201 & 0.309 &  0.440 
       & 0.405 & 0.377 & 0.412 & 0.453
       & 0.361 & 0.260 & 0.396 & 0.559 \\
\hline
\multicolumn{13}{l}{\emph{Low-dimensional Methods}} \\
\hline
MuRP \citep{balazevic2019multi} & 0.323 & 0.235 & 0.353 & \underline{0.501} 
       & 0.465 & 0.420 & 0.484 & 0.544 
       & 0.230 & 0.150 & 0.247 & 0.392 \\
AttH   \citep{chami2020low} & 0.324 & 0.236 & 0.354 & \underline{0.501} 
       & \underline{0.466} & \underline{0.419} & \underline{0.484} & \underline{0.551} 
       & 0.397 & 0.310 & 0.437 & 0.566 \\
DualDE \citep{zhu2022dualde} & 0.306 & 0.216 & 0.338 & 0.489 
       & \textbf{0.468} & \textbf{0.419} & \textbf{0.486} & \textbf{0.560} 
       & - & - & - & - \\
\hline\hline
TransE + GreenKGC (Ours) & \underline{0.331} & \underline{0.251} & \underline{0.356} & 0.493
       & 0.342 & 0.300 & 0.365 & 0.413
       & 0.362 & 0.265& 0.408 & 0.537 \\
RotatE + GreenKGC (Ours) & \textbf{0.345} & \textbf{0.265} & \textbf{0.369} & \textbf{0.507}
       & 0.411 & 0.367 & 0.430 & 0.491
       & \textbf{0.453} & \textbf{0.361} & \textbf{0.509} & \textbf{0.629} \\
\hline
\end{tabular}
}
\caption{Results of link prediction in low dimensions ($d=32$), where the best
and the second best numbers are in bold and with an underbar, respectively.}
\label{tab:low}
\end{table*}

\subsection{Experimental Setup}

{\bf Datasets.} We consider four link prediction datasets for 
performance benchmarking: FB15k-237~\citep{bordes2013translating,
toutanova2015observed}, WN18RR~\citep{bordes2013translating,
dettmers2018convolutional}, YAGO3-10~\citep{dettmers2018convolutional},
and ogbl-wikikg2 \citep{hu2020open}.
Their statistics are summarized in Table \ref{table:dataset}. 
FB15k-237 is a subset of Freebase
\citep{bollacker2008freebase} that contains real-world relationships.
WN18RR is a subset of WordNet \citep{miller1995wordnet} containing
lexical relationships between word senses. YAGO3-10 is a subset of YAGO3
\citep{mahdisoltani2014yago3} that describes the attributes of persons.
ogbl-wikikg2 is extracted from wikidata \citep{vrandevcic2014wikidata}
capturing the different types of relations between entities in the world.
Among the four, ogbl-wikikg2 is a large-scale dataset with more than 2.5M
entities.

{\bf Implementation details.} We adopt TransE \citep{bordes2013translating} 
and RotatE \citep{sun2018rotate} as the baseline models and learn 500 
dimensions initial representations for entities and relations. The feature
dimensions are then reduced in the feature pruning process.
We compare among GreenKGC using RotatE as the baseline in all ablation studies.
To partition the KG, we determine the number of groups $k$ for each dataset
when the average cross-entropy of all feature dimensions converges. As a result,
$k=3$ for WN18RR, $k=5$ for FB15k-237 and YAGO3-10, and $k=20$ for ogbl-wikikg2.

For decision learning, we consider several tree-based binary classifiers, 
including Decision Trees~\citep{breiman2017classification}, 
Random Forest~\citep{breiman2001random}, and Gradient Boosting Machines 
\citep{chen2016xgboost}, as they match the intuition of the feature 
pruning process and can accommodate non-linearity in the sample
distribution. The hyperparameters are searched among: 
tree depth $l \in$ \{3, 5, 7\}, number of estimators $n \in$ \{400, 800,
1,200, 1,600, 2,000\}, and learning rate $lr \in$ \{0.05, 0.1, 
0.2\}. The best settings are chosen based on MRR in the validation set. As 
a result, we adopt Gradient Boosting Machine for all datasets. 
$l = 5$, $n = 1200$, $lr = 0.2$ for FB15k-237 and YAGO3-10, 
$l = 3$, $n = 1600$, $lr = 0.1$ for WN18RR, and $l = 7$, $n = 2000$, 
$lr = 0.05$ for ogbl-wikikg2. We adopt ontology-based
negative sampling to train classifiers for FB15k-237, YAGO3-10, 
and ogbl-wikikg2, and embedding-based negative sampling for WN18RR. 
Baseline KGEs are trained on NVIDIA Tesla P100 GPUs and binary 
classifiers are trained on AMD EPYC 7542 CPUs.

{\bf Evaluation metrics.} For the link prediction task, the goal is to
predict the missing entity given a query triple, i.e. $(h, r, ?)$ or
$(?, r, t)$. The correct entity should be ranked higher than other
candidates. Here, several common ranking metrics are used, such as MRR
(Mean Reciprocal Rank) and Hits@k (k=1, 3, 10). Following the convention
in~\citet{bordes2013translating}, we adopt the filtered setting, where
all entities serve as candidates except for the ones that have been seen
in training, validation, or testing sets.

\begin{table*}[!t]
\setlength\tabcolsep{3pt}
\centering
\begin{tabular}{l|c | ccc | ccc | ccc}
\hline
&
& \multicolumn{3}{c|}{\textbf{FB15k-237}} 
& \multicolumn{3}{c|}{\textbf{WN18RR}}
& \multicolumn{3}{c}{\textbf{YAGO3-10}} \\
\hline
Baseline & Dim. & MRR & H@1 & \#P (M) 
& MRR & H@1 & \#P (M) 
& MRR & H@1 & \#P (M) \\
\hline 
\multirow{5}{*}{TransE} &
500
& 0.325 & 0.228 & 7.40 
& 0.223 & 0.013 & 20.50
& 0.416 & 0.319 & 61.60 \\

\cline{2-11}

&\multirow{2}{*}{100} 
& 0.274 & 0.186 & 1.48 
& 0.200 & 0.009 & 4.10
& 0.377 & 0.269 & 12.32 \\
&& \textcolor{red}{$\downarrow$} 15.7\% & \textcolor{red}{$\downarrow$} 18.5\% & (0.20x) 
& \textcolor{red}{$\downarrow$} 10.3\% & \textcolor{red}{$\downarrow$} 30.8\% & (0.20x) 
& \textcolor{red}{$\downarrow$} 9.4\% & \textcolor{red}{$\downarrow$} 16.7\% & (0.20x) \\

\cline{2-11}

& 100
& 0.338 & 0.253 & 1.76 
& 0.407 & 0.361 & 4.38 
& 0.455 & 0.358 & 12.60 \\
& (Ours)
& \textcolor{green}{$\uparrow$} 4.0\% & \textcolor{green}{$\uparrow$} 9.6\% & (0.24x)
& \textcolor{green}{$\uparrow$} 82.5\% & \textcolor{green}{$\uparrow$} 176.9\% & (0.21x) 
& \textcolor{green}{$\uparrow$} 9.4\% & \textcolor{green}{$\uparrow$} 12.2\% & (0.20x) \\
\hline\hline

\multirow{5}{*}{RotatE} &
500
& 0.333 & 0.237 & 14.66 
& 0.475 & 0.427 & 40.95
& 0.478 & 0.388 & 123.20 \\

\cline{2-11}

&\multirow{2}{*}{100} 
& 0.296 & 0.207 & 2.93 
& 0.437 & 0.385 & 8.19
& 0.432 & 0.340 & 24.64 \\
&& \textcolor{red}{$\downarrow$} 11.1\% & \textcolor{red}{$\downarrow$} 12.7\% & (0.20x)
& \textcolor{red}{$\downarrow$} 8\% & \textcolor{red}{$\downarrow$} 9.8\% & (0.20x)
& \textcolor{red}{$\downarrow$} 9.6\% & \textcolor{red}{$\downarrow$} 12.4\% & (0.20x) \\

\cline{2-11}

& 100
& 0.348 & 0.266 & 3.21 
& 0.458 & 0.424 & 8.47 
& 0.467 & 0.378 & 24.92 \\
& (Ours)
& \textcolor{green}{$\uparrow$} 4.5\% & \textcolor{green}{$\uparrow$} 12.2\% & (0.22x)
& \textcolor{red}{$\downarrow$} 3.6\% & \textcolor{red}{$\downarrow$} 0.7\% & (0.21x) 
& \textcolor{red}{$\downarrow$} 2.3\% & \textcolor{red}{$\downarrow$} 3.6\% & (0.20x) \\
\hline
\end{tabular}
\caption{Results on the link prediction task, where we show the
performance gain (or loss) in terms of percentages with an up (or down)
arrow and the ratio of the model size within the parentheses against
those of respective 500-dimensional models.} \label{tab:main}
\end{table*}

\begin{figure}[t]
\centering
\includegraphics[width=0.6\textwidth]{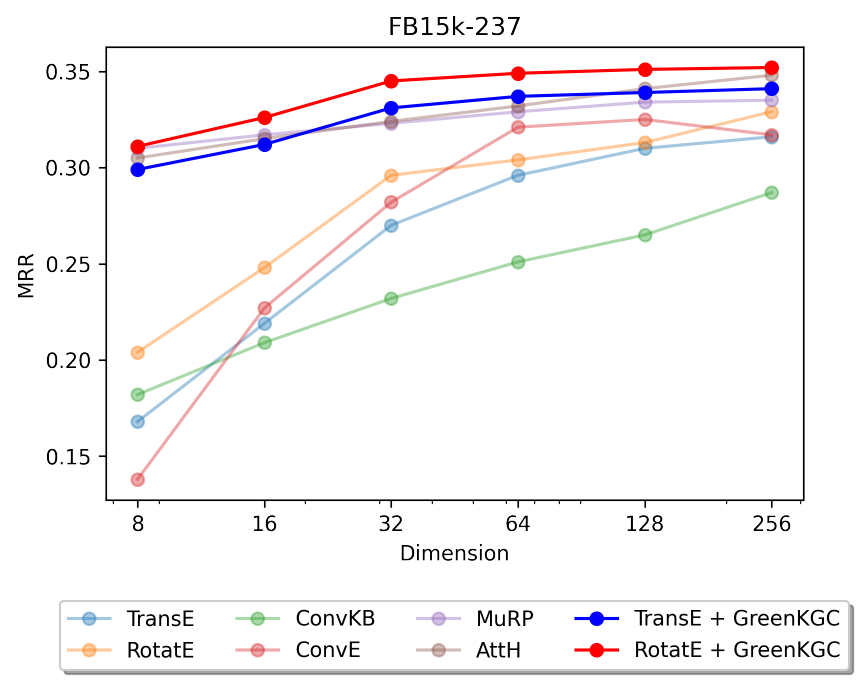}
\caption{Embedding dimension $d$ to MRR curves in log-scale for various 
methods on FB15k-237. $d$ = 8, 16, 32, 64, 128, 256.}\label{fig:dim_curve}
\end{figure}

\subsection{Main Results}

\textbf{Results in low dimensions.} In Table \ref{tab:low}, we compare
GreenKGC with KGE, classification-based, and low-dimensional KGE methods in
low dimensions, i.e. $d$ = 32. Results for other methods in Table \ref{tab:low} 
are either directly taken from \citep{chami2020low, zhu2022dualde}
or, if not presented, trained by ourselves using publicly available 
implementations with hyperparameters suggested by the original papers.
KGE methods cannot achieve good performance in low dimensions 
due to over-simplified scoring functions. Classification-based methods achieve 
performance better than KGE methods as they adopt NNs as complex
decoders. Low-dimensional KGE methods provide state-of-the-art 
KGC solutions in low dimensions. Yet, GreenKGC outperforms them 
in FB15k-237 and YAGO3-10 in all metrics.
{\color{black}
For WN18RR, the baseline KGE methods perform poorly in low dimensions.
GreenKGC is built upon
KGEs, so this affects the performance of GreenKGC in WN18RR. 
Thus, GreenKGC is more suitable for instance-based KGs, such as 
Freebase and YAGO, while hyperbolic KGEs, such as MuRP and AttH model
the concept-based KGs, such as WordNet, well.
}

We show the performance curves of various methods as a function of
embedding dimensions in Fig. \ref{fig:dim_curve}. We see that the
performance of KGE methods (i.e. TransE and RotatE) drops significantly 
as the embedding dimension is lower. For ConvKB, although its
performance is less influenced by dimensions due to a complex decoder, 
it performs poorly compared to other methods in general. For ConvE, although
it claims it's more efficient in parameter scaling 
\citep{dettmers2018convolutional}, its performance actually degrades 
significantly in dimensions lower than 64. In addition, it also doesn't
perform well when the dimension is larger. Thus, the performance of ConvE is 
sensitive to the embedding dimension. 
MuRP, AttH, and GreenKGC are the only methods that can offer 
reasonable performance as the dimension goes to as low as 8 dimensions. 

\begin{table}[t]
\color{black}
\setlength\tabcolsep{3pt}
\centering
\begin{tabular}{l|r|c|c}
\hline
Method & \#P (M) & Val. MRR & Test MRR  \\
\hline
TransE (d = 500) & 1,250 (5$\times$) & 0.427 & 0.426  \\
RotatE (d = 250) & 1,250 (5$\times$) & 0.435 & 0.433  \\
\hline
TransE (d = 100) & 250 (1$\times$) & 0.247 & 0.262  \\
TransE + GreenKGC (d = 100) & 250 (1$\times$) & \textbf{0.339} & \textbf{0.331}  \\\hline
RotatE (d = 50)  & 250 (1$\times$) & 0.225 & 0.253  \\
RotatE + GreenKGC (d = 50)  & 250 (1$\times$) & \textbf{0.341} & \textbf{0.336} \\
\hline
\end{tabular}
\caption{Link prediction performance on obgl-wikikg2 dataset. }
\label{tab:ogbl}
\end{table}

\textbf{Comparison with baseline KGE.} One unique characteristic of
GreenKGC is to prune a high-dimensional KGE into low-dimensional triple
features and make predictions with a binary classifier as a powerful
decoder. We evaluate the capability of GreenKGC in saving the number of
parameters and maintaining the performance by pruning original 
500-dimensional KGE to 100-dimensional triple features in Table \ref{tab:main}. 
As shown in the table, GreenKGC can achieve competitive or even better 
performance with around 5 times smaller model size. Especially, Hits@1 is 
retained the most and even improved compared to the high-dimensional baselines. 
In addition, GreenKGC using TransE as the baseline can outperform 
high-dimensional TransE in all datasets. Since the 
TransE scoring function is simple and fails to model some relation patterns, such 
as symmetric relations, incorporating TransE with a powerful decoder, i.e. a 
binary classifier, in GreenKGC successfully overcomes deficiencies of adopting 
an over-simplified scoring function. 
For all datasets, 100-dimensional GreenKGC could generate better results than
100-dimensional baseline models.

{\color{black}We further compare GreenKGC and its baseline KGEs on a large-scale 
dataset, ogbl-wikikg2. Table \ref{tab:ogbl} shows the results. We reduce the feature 
dimensions from 500 to 100 for RotatE and 250 to 50 for TransE and achieve a 5x smaller
model size while retaining around 80\% of the performance. Compared with the baseline 
KGEs in the same feature dimension, GreenKGC can improve 51.6\% in MRR for RotatE and 
37.2\% in MRR for TransE. Therefore, the results demonstrate the advantages in performance
to apply GreenKGC to large-scale KGs in a constrained resource.}

\subsection{Ablation Study} \label{sec:ablation} 

\begin{table}[t]
\setlength\tabcolsep{3pt}
\centering
{\color{black}
\begin{tabular}{l | c  c  c | c  c  c}
\hline
& \multicolumn{3}{c|}{\textbf{FB15k-237}} & \multicolumn{3}{c}{\textbf{WN18RR}} \\
\hline
& MRR & H@1 & H@10 & MRR & H@1 & H@10\\
\hline
w/o pruning   & 0.318 & 0.243 & 0.462 & 0.379 & 0.346 & 0.448 \\
\hline
random             & 0.313 & 0.239 & 0.460 & 0.375 & 0.346 & 0.420 \\
variance           & 0.315 & 0.239 & 0.465 & 0.381 & 0.348 & 0.455 \\
feature importance & 0.323 & 0.241 & 0.478 & 0.385 & 0.355 & 0.464 \\
\hline
prune low CE             & 0.312 & 0.236 & 0.460 & 0.373 & 0.343 & 0.419 \\
prune high CE (Ours)     & {\bf 0.345} & {\bf 0.265} & {\bf 0.507} & 
               {\bf 0.411} & {\bf 0.367} & {\bf 0.491} \\
\hline
\end{tabular}
}
\caption{Performance for RotatE + GreenKGC in 32
dimensions with different feature pruning scheme.}\label{tab:prune}
\end{table}

\begin{figure*}[t]
\color{black}
        \begin{subfigure}[b]{0.32\textwidth}
        \centering
         \includegraphics[width=\textwidth]{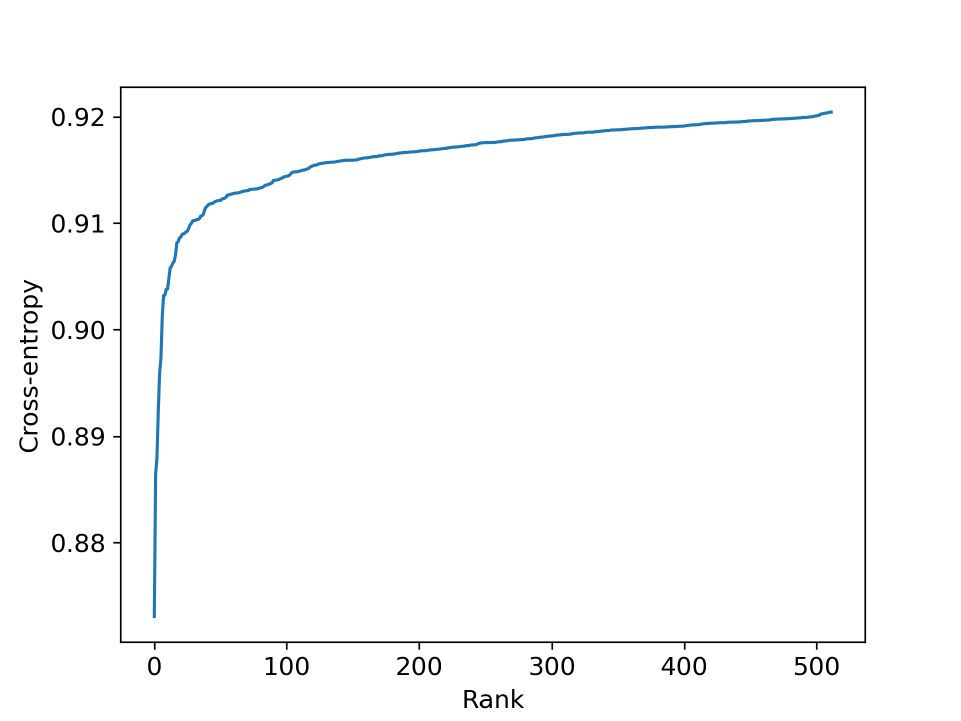}
        \caption{Cross-entropy in DFT}
        \end{subfigure}
        \hfill
        \begin{subfigure}[b]{0.32\textwidth}
        \centering
         \includegraphics[width=\textwidth]{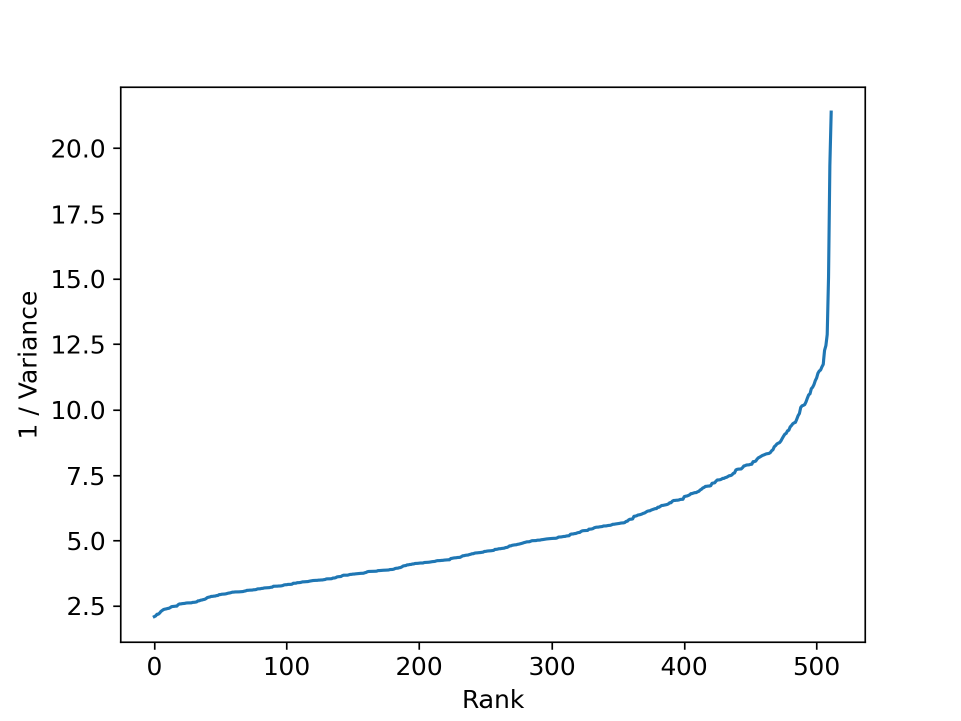}
        \caption{1 / Variance}
        \end{subfigure}
        \hfill
        \begin{subfigure}[b]{0.32\textwidth}
        \centering
         \includegraphics[width=\textwidth]{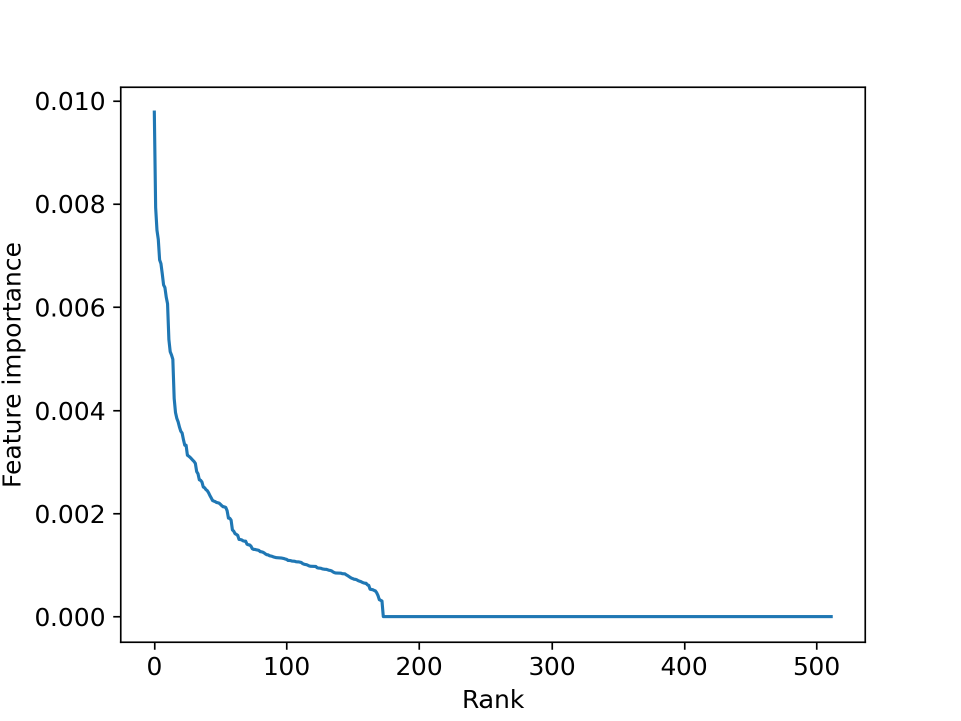}
        \caption{Feature importance}
        \end{subfigure}
        \hfill
\caption{\textcolor{black}{Sorted discriminability for each feature dimension in different feature pruning schemes. For cross-entropy and 1/variance, a lower value indicates a more discriminant feature. For feature importance, a higher value indicates a more discriminant feature.}}
\label{fig:dft}
\end{figure*}

\textbf{Feature pruning.} We evaluate the effectiveness
of the feature pruning scheme in GreenKGC in Table \ref{tab:prune}. We
use ``w/o pruning" to denote the baseline 32 dimensions KGE directly 
followed by the decision learning module. 
{\color{black}
Also, we compare the following feature pruning schemes: 1) random pruning, 
2) pruning based on variance, 3) pruning based on 
feature importance from a Random Forest classifier, 4) pruning dimensions 
with low CE (i.e. the most discriminant ones), in DFT, and 5) pruning 
dimensions with high CE (i.e. the least discriminant ones) in DFT.
As shown in the table, our method to prune the least discriminant features in
DFT achieves the best performance on both
datasets. In contrast, pruning the most discriminant features in DFT performs the worst. Thus, DFT module can effectively differentiate the discriminability among different features.
Using variance to prune achieves similar results as ``w/o pruning" and random pruning.
Pruning based on feature importance shows better results than ``w/o pruning", random 
and pruning, and pruning based on variance, but performs worse than DFT. In addition, feature
importance needs to consider all feature dimensions at once, while in DFT, each feature 
dimension is processed individually. Thus, DFT is also more memory efficient than calculating
feature importance.

Fig. \ref{fig:dft} plots the sorted discriminability of features in different pruning 
schemes. From the figure, the high variance region is flat, so it's difficult to identify the
most discriminant features using their variances. For feature importance, some of the feature
dimensions have zero scores. Therefore, pruning based on feature importance
might ignore some discriminant features. In the DFT curve, there is a ``shoulder point"
indicating only around 100 feature dimensions are more discriminant than the others. 
In general, we can get good performance in low dimensions as long as we preserve 
dimensions lower than the shoulder point and prune all other dimensions.
}
\begin{figure}[t]
\centering
     \begin{subfigure}[b]{0.48\textwidth}
         \centering
         \includegraphics[width=\textwidth]{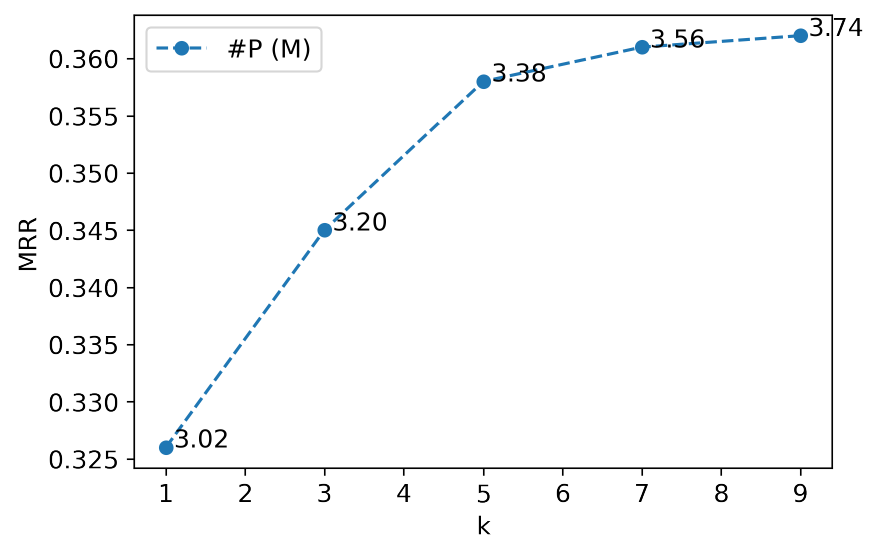}
         \caption{FB15k-237}
     \end{subfigure}
     \begin{subfigure}[b]{0.48\textwidth}
         \centering
         \includegraphics[width=\textwidth]{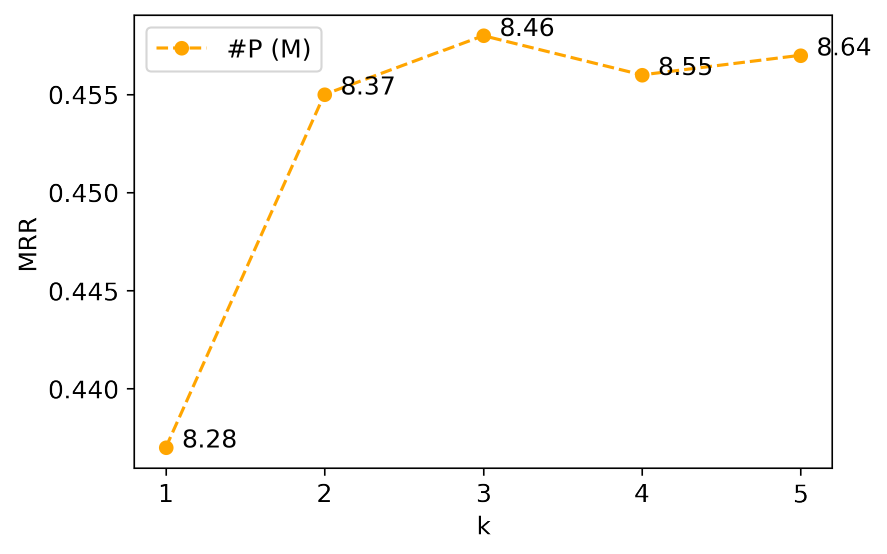}
         \caption{WN18RR}
     \end{subfigure}
\caption{\color{black}Ablation study on number of relation groups $k$ to MRR.}\label{fig:partition}
\end{figure}

{\color{black}
\textbf{KG partitioning.} Figure \ref{fig:partition} shows GreenKGC performance with
different numbers of relation groups $k$, where $k = 1$ means no KG partitioning.
A larger $k$ will give a better performance on both FB15k-237 and WN18RR. Without using KG
partitioning performs much worse than using KG partitioning.
Note that with a larger $k$, GreenKGC has more model parameters since we need more classifiers.
The model complexity is $O(|E|d + k\Theta)$, where $\Theta$ is the model complexity for the
classifier. Thus, we can adjust $k$ based on the tradeoff of performance convergence 
and memory efficiency.
}

\section{Conclusion and Future Work} \label{sec:conclusion}

A lightweight KGC method, called GreenKGC, was proposed in this work
to make accurate link predictions in low dimensions. It consists of three modules that
can be trained individually: 1) representation learning, 2) feature pruning,
and 3) decision learning. Experimental results in low dimensions demonstrate
GreenKGC can achieve satisfactory performance in as low as 8 dimensions. In addition,
experiments on ogbl-wikikg2 show GreenKGC can get competitive results 
with much fewer model parameters. Furthermore, the ablation study shows the 
effectiveness of KG partitioning and feature pruning.

{\color{black} Modularized GreenKGC allows several future extensions. 
First, GreenKGC can be combined with new embedding
models as initial features. In general, using a more expressive KGE model can
lead to better final performance. Second, individual modules can
be fine-tuned for different applications. For example, since the feature
pruning module and the decision-learning module are supervised, they can
be applied to various applications. Finally, different negative sampling
strategies can be investigated in different applications.}

\bibliographystyle{plainnat}
\renewcommand\refname{Reference}
\bibliography{main}

\newpage
\section{Appendix}

\subsection{Training Procedure for Baseline KGE Models}\label{appendix:emb}

\begin{table}[ht]
\setlength\tabcolsep{3pt}
\centering
\begin{tabular}{ l  c  c  c  c }
\toprule
Model & $n_e$ & $n_r$ & $n_v$ & $f_r(h, t)$\\
\midrule
TransE   & 1 & 1 & 3 & $-\| \bm{h} + \bm{r} - \bm{t} \|$ \\
DistMult & 1 & 1 & 3 & $\langle \bm{h}, \bm{r}, \bm{t} \rangle$ \\
ComplEx  & 2 & 2 & 6 & $Re(\langle \bm{h}, \bm{r}, \overline{\bm{t}} \rangle)$ \\
RotatE   & 2 & 1 & 5 & $-\| \bm{h} \circ \bm{r} - \bm{t} \|^2$ \\
\bottomrule
\end{tabular}
\caption{Popular KGE methods and their scoring functions, where $\bm{h}$, $\bm{r}$, 
and $\bm{t}$ denote embeddings for a given triple $(h, r, t)$, $d$ is the
embedding dimension. $ \circ $ denotes the Hadamard product, and
$\langle \cdot, \cdot, \cdot \rangle$ is the generalized dot product. $n_e$ 
is the number of entity variables in one dimension, $n_r$ 
is the number of relation variables in one dimension, and $n_v$ 
is the number of triple variables in one dimension. $n_v = 2 n_e + n_r$.}
\label{tab:models}
\end{table}

To train the baseline KGE model as the initial entity and relation
representations, we adopt the self-adversarial learning process in
\citet{sun2018rotate} and use this 
codebase\footnote{\url{https://github.com/DeepGraphLearning/KnowledgeGraphEmbedding}}. 
That is, given an observed triple $(h, r, t)$ and
the KGE model $f_r(\bm{h}, \bm{t})$, we minimize the following loss function
\begin{equation}
\begin{split}
        \mathcal{L} = {} & - \log(\sigma(f_r(\bm{h}, \bm{t}))) \\
                         & - \sum_{i=1}^n p(h'_i, r, t'_i) 
                    \log(\sigma( - f_r(\bm{h}'_i, \bm{t}'_i))), 
\end{split}
\end{equation}
where $(h'_i, r, t'_i)$ is a negative sample and 
\begin{equation}
        p(h'_j, r, t'_j) = \frac{\exp ( \alpha f_r(\bm{h}'_j, \bm{t}'_j) )}
{\sum_{i=1}^n \exp ( \alpha f_r(\bm{h}'_i, \bm{t}'_i) )},
\end{equation}
where $\alpha$ is the temperature to control the self-adversarial 
negative sampling. We summarize the scoring functions for 
some common KGE models and their corresponding number of variables per
dimension in Table \ref{tab:models}. In general, 
GreenKGC can build upon any existing KGE models.

\subsection{DFT Implementation Details}\label{appendix:dft}

To calculate the discriminant power of each dimension, we iterate
through each dimension in the high-dimension feature set and calculate
the discriminant power based on sample labels. More specifically,
we model KGC as a binary classification task. We assign label $y_i = 1$
to the $i$th sample if it is an observed triple and $y_i = 0$ if it
is a negative sample. For the $d$th dimension, we split the 1D
feature space into left and right subspaces and calculate the cross-entropy
in the form of
\begin{equation}
        H^{(d)} = \frac{N_LH_L^{(d)} + N_RH_R^{(d)}}{N_L + N_R},
\end{equation}
where $N_L$ and $N_R$ are the numbers of samples in the left and right intervals,
respectively,
\begin{eqnarray}
        H_L^{(d)} = - P_{L, 1}\log(P_{L, 1}) - P_{L, 0}\log(P_{L, 0}), \\
        H_R^{(d)} = - P_{R, 1}\log(P_{R, 1}) - P_{R, 0}\log(P_{R, 0}),
\end{eqnarray}
and where $P_{L, 1} = \frac{1}{N_L} \sum_{i=1}^{N_L} y_i$, and 
$P_{L, 0} = 1 - P_{L, 1}$ and similarly for $P_{R, 1}$ and $P_{R, 0}$. A
lower cross-entropy value implies higher discriminant power. 

\begin{figure}[t]
\centering
     \begin{subfigure}[b]{0.48\textwidth}
         \centering
         \includegraphics[width=\textwidth]{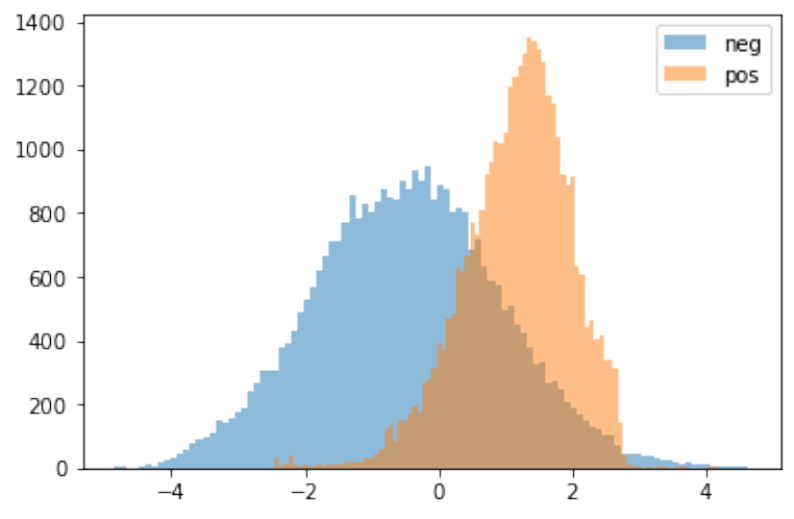}
         \caption{Cross-entropy $= 0.7348$}
     \end{subfigure}
     \begin{subfigure}[b]{0.48\textwidth}
         \centering
         \includegraphics[width=\textwidth]{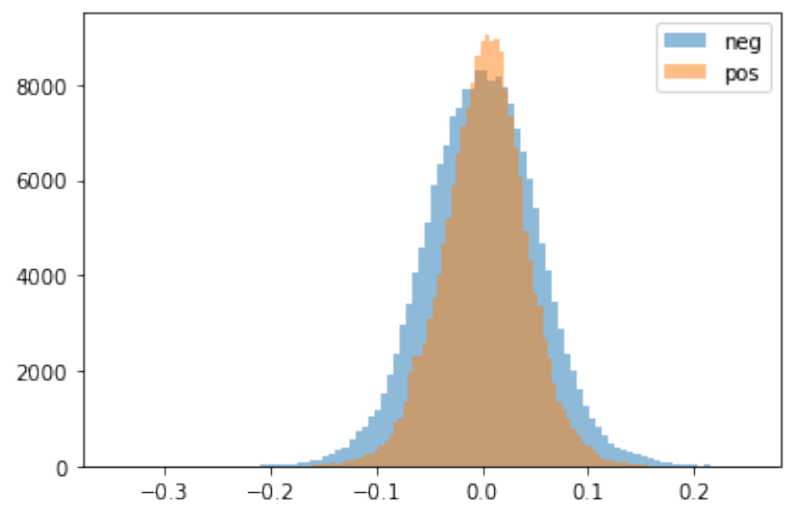}
         \caption{Cross-entropy $= 0.9910$}
     \end{subfigure}
\caption{Histograms of PCA-transformed 1D triple variables in two feature dimensions with (a) low and (b) high cross-entropy.}\label{fig:dim_hist}
\end{figure}

Fig. \ref{fig:dim_hist} shows histograms of linearly transformed 1D triple 
variables in two different feature dimensions. As seen in the figure, samples in
Fig. \ref{fig:dim_hist} (a), i.e. the feature dimension with the lower 
cross-entropy, are more separable than that in Fig. \ref{fig:dim_hist} (b), i.e.
the feature dimension with the higher cross-entropy. Therefore, a lower 
cross-entropy implies a more discriminant feature dimension.




\begin{table*}[!t]
\color{black}
\setlength\tabcolsep{3pt}
\centering
\begin{tabular}{l | c  c  c  c | c  c  c  c }
\hline
& \multicolumn{4}{c|}{\textbf{Predicting Heads}} & \multicolumn{4}{c}{\textbf{Predicting Tails}} \\
\hline
Model & 1-to-1 & 1-to-N & N-to-1 & N-to-N & 1-to-1 & 1-to-N & N-to-1 & N-to-N \\
\hline
TransE \cite{bordes2013translating}  &0.374 &0.417 &0.037 &0.217 &0.372 &0.023 &0.680 &0.322\\
RotatE \cite{sun2018rotate}          &0.468 &0.431 &0.066 &0.229 &0.463 &0.057 &0.725 &0.336\\
AttH \cite{chami2020low}             &0.473 &0.432 &0.071 &0.236 &0.472 &0.057 &0.728 &0.343\\ 
\hline
TransE + GreenKGC (Ours)     &0.478 &0.442 &0.088 &0.243 &0.477 &0.096 &0.754 &0.351\\
RotatE + GreenKGC (Ours)     &\textbf{0.483} &\textbf{0.455} &\textbf{0.134} &\textbf{0.245} &\textbf{0.486} &\textbf{0.112} &\textbf{0.765} &\textbf{0.353}\\
\hline
\end{tabular}
\caption{Performance on different relation categories in FB15k-237 under 32 dimensions.}
\label{tab:rel}
\end{table*}

{\color{black}
\subsection{Relation Categories}
We further evaluate GreenKGC in different relation categories. Following the 
convention in \citet{wang2014knowledge}, we divide the relations into four
categories: 1-to-1, 1-to-N, N-to-1, and N-to-N. They are characterized by two
statistical numbers, head-per-tail (hpt), and tail-per-head (tph), of the datasets.
If $tph < 1.5$ and $hpt < 1.5$, the relation is treated as 1-to-1;
if $tph < 1.5$ and $hpt \geq 1.5$, the relation is treated as 1-to-N;
if $tph \geq 1.5$ and $hpt < 1.5$, the relation is treated as N-to-1;
if $tph \geq 1.5$ and $hpt \geq 1.5$, the relation is treated as N-to-N.

Table \ref{tab:rel} summarizes the results for different relation categories in
FB15k-237 under 32 dimensions. In the low-dimensional setting, GreenKGC is able to outperform
other methods in all relation categories. Specifically, GreenKGC performs especially 
well for many-to-1 predictions (i.e. predicting heads for 1-to-N relations, and predicting
tails for N-to-1 relations). Such results demonstrate the advantage of using classifiers 
to make accurate predictions when there is only one valid target.
}

\begin{table}[t]
\setlength\tabcolsep{3pt}
\centering
\begin{tabular}{l | c | c | c}
\hline
& \textbf{FB15k-237} & \textbf{WN18RR} & \textbf{YAGO3-10} \\
\hline
DualDE   & 03:30:50 & 01:50:00 & 09:28:20 \\
GreenKGC (Ours) & 00:10:50 & 00:06:02 & 00:23:35 \\
\hline
\end{tabular}
\caption{Comparison of required training time (Hour : Minute : Second) to 
reduce the feature dimensions from 512 to 100 for TransE 
between DualDE, a knowledge-distillation method, and GreenKGC.}
\label{tab:reduction_time}
\end{table}

\begin{table*}[t]
\setlength\tabcolsep{3pt}
\centering
\resizebox*{\textwidth}{!}{
\begin{tabular}{l | c  c  c  c  c  c | c  c  c  c  c  c}
\hline
& \multicolumn{6}{c|}{\textbf{FB15k-237}} & \multicolumn{6}{c}{\textbf{WN18RR}} \\
\hline
Model & MRR & H@1 & H@3 & H@10 & \#P (M) & T (s) & MRR & H@1 & H@3 & H@10 & \#P (M) & T (s) \\
\hline
ConvKB \citep{nguyen2018novel} & 0.258 & 0.179 & 0.283 & 0.416 & 1.91 & 548.67
       & 0.369 & 0.317 & 0.399 & 0.468 & 5.26 & 225.12 \\
ConvE \citep{dettmers2018convolutional} & 0.317 & 0.230 & 0.347 & 0.493 & 2.74 & 235.73
       & 0.427 & \textbf{0.394} & 0.437 & 0.495 & 6.09 & 46.08 \\
\hline
TransE + GreenKGC (Ours) & \textbf{0.339} & \textbf{0.253} & \textbf{0.364} & \textbf{0.503} & 2.42 & 205.12 
& \textbf{0.435} & 0.391 & \textbf{0.461} & \textbf{0.510} & 5.84 & 40.01 \\
\hline
\end{tabular}
}
\caption{Comparison on performance, number of model parameters, and
total inference time (batch size = 8) with other classification-based methods in 128
dimensions. We adopt TransE as the baseline for fair 
comparison in the number of model parameters. The best numbers are in bold.} 
\label{tab:cls}
\end{table*}

\begin{figure}[t]
\centering
\includegraphics[width=0.5\textwidth]{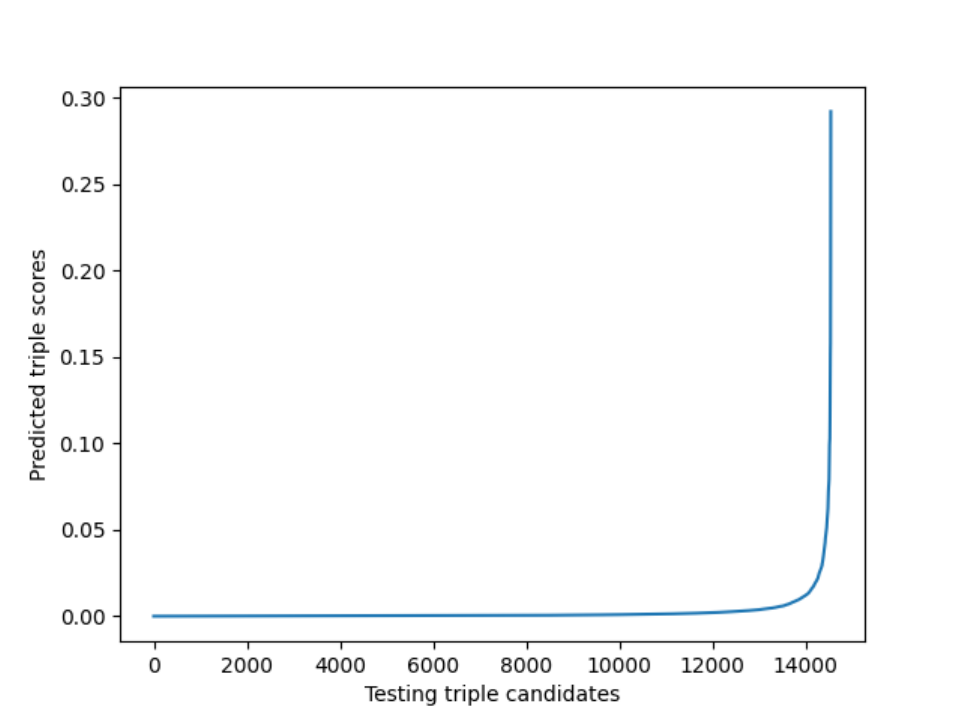}
\caption{Prediction distribution of a query (\emph{38th Grammy Awards}, \emph{award\_winner}, \emph{?}) in FB15k-237. A higher predicted score implies a higher chance of being a valid triple. }
\label{fig:distribution}
\end{figure}

\begin{table}[t]
\setlength\tabcolsep{3pt}
\centering
\begin{tabular}{l | c  c  c | c  c  c}
\hline
& \multicolumn{3}{c|}{\textbf{FB15k-237}} & \multicolumn{3}{c}{\textbf{WN18RR}} \\
\hline
Neg. sampling & MRR & H@1 & H@10 & MRR & H@1 & H@10\\
\hline
Random        & 0.283 & 0.197 & 0.452 & 0.407 & 0.361 & 0.481 \\
Ontology      & {\bf 0.345} & {\bf 0.265} & {\bf 0.507} & 0.403 & 0.350 & 0.487 \\
Embedding     & 0.316 & 0.232 & 0.471 & {\bf 0.411} & {\bf 0.367} & {\bf 0.491} \\
\hline
\end{tabular}
\caption{Ablation study on different negative sampling methods for classifier 
training in 32 dimensions.}\label{tab:neg}
\end{table}

\begin{figure}[t]
\centering
     \begin{subfigure}[b]{0.48\textwidth}
         \centering
         \includegraphics[width=\textwidth]{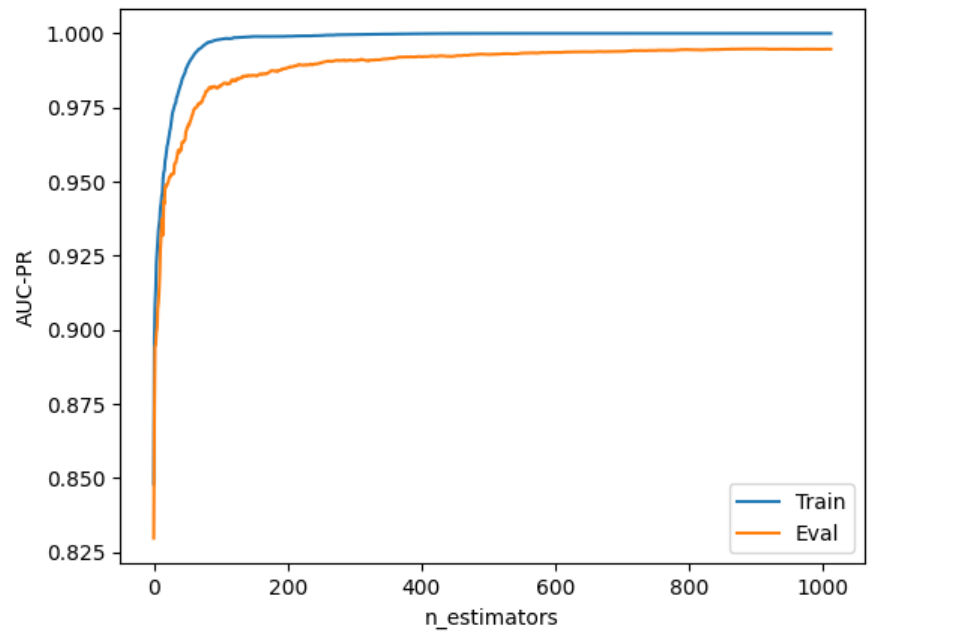}
         \caption{Training/evaluation AUC-PR.} \label{fig:aucpr}
     \end{subfigure}
     \begin{subfigure}[b]{0.48\textwidth}
         \centering
         \includegraphics[width=\textwidth]{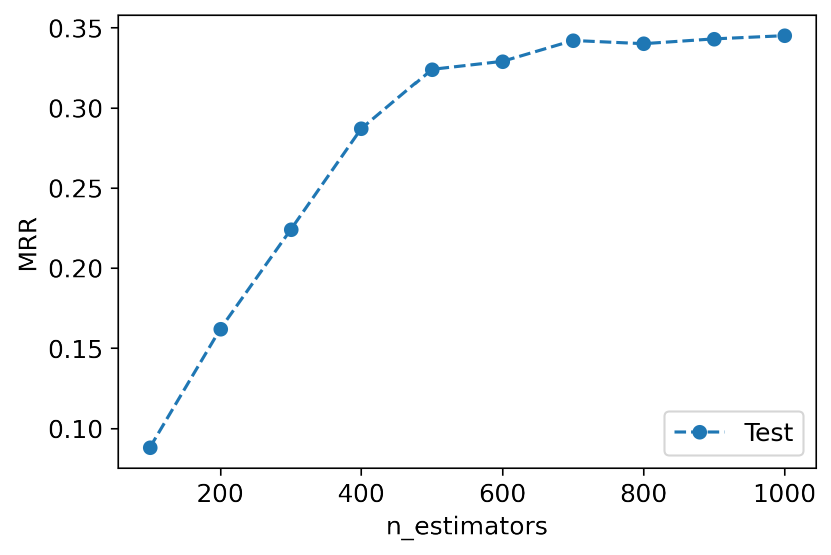}
         \caption{Testing MRR.} \label{fig:mrr}
     \end{subfigure}
\caption{\color{black} Training/evaluation AUC-PR and testing MRR to the number of training iterations.}
\label{fig:training}
\end{figure}

\subsection{Time Analysis on Feature Pruning} Table 
\ref{tab:reduction_time} shows the required training time for DualDE 
\citep{zhu2022dualde}, a knowledge distillation method, and GreenKGC, to reduce 
512 dimensions TransE embeddings to 100 dimensions. As shown in the table, GreenKGC 
achieves around 20x faster training time compared to DualDE, 
especially in YAGO3-10, which is a larger-scale dataset. Besides, in knowledge 
distillation methods, low-dimensional embeddings are randomly initialized and 
trained with the guidance of high-dimensional embeddings. Thus, the quality of 
the low-dimensional embeddings highly depends on good initialization. On the 
contrary, the feature pruning process in GreenKGC selects a subset of powerful 
feature dimensions without learning new features from scratch. In addition, it 
is also memory-efficient since it only processes one feature dimension at 
once.

\subsection{Comparison with NN-based Methods}

\textbf{Inference time analysis.} We compare GreenKGC with two other 
NN-based methods in Table \ref{tab:cls} in terms of performance, 
number of free parameters, and inference time. They are ConvKB \citep{nguyen2018novel}
and ConvE \citep{dettmers2018convolutional}. We adopt TransE as the baseline in GreenKGC
to match the number of parameters in the embedding layer for a fair comparison.
As compared with ConvKB, GreenKGC achieves significantly better performance
with slightly more parameters. As compared with ConvE, GreenKGC uses fewer 
parameters and demands a shorter inference time since ConvE adopts a multi-layer 
architecture. GreenKGC also offers better performance compared to ConvE.

\textbf{Prediction distribution.} It was reported in \citet{sun2020evaluation} 
that the predicted scores for all candidates on FB15k-237 are converged to 1 with ConvKB
\citep{nguyen2018novel}.  This is unlikely to be true, given the fact
that KGs are often highly sparse. The issue is resolved after ConvKB is
implemented with PyTorch\footnote{\url{https://github.com/daiquocnguyen/ConvKB/issues/5}}, 
but the performance on FB15k-237 is still not as good as ConvKB
originally reported in the paper. The issue shows the problem of
end-to-end optimization. That is, it is difficult to control and monitor
every component in the model. This urges us to examine whether GreenKGC
has the same issue. Fig. \ref{fig:distribution} shows the sorted
predicted scores of a query (\emph{38th Grammy Awards}, \emph{award\_winner},
\emph{?}) in FB15k-237. We see from the figure that
only very few candidates have positive scores close to $1$, while other
candidates receive negative scores of 0. The formers are valid triples.
The score distribution is consistent with the sparse nature of KGs.

\subsection{Ablation on Negative Sampling}

We evaluate the 
effectiveness of the two proposed negative sampling
(i.e., ontology- and embedding-based) methods in Table \ref{tab:neg}. In
FB15k-237, both are more effective than randomly drawn negative samples.
The ontology-based one gives better results than the embedding-based
one. In WN18RR, the embedding-based one achieves the best results.
Since there is no clear entity typing in WordNet, the
ontology-based one performs worse than the randomly drawn one. We can
conclude that to correct failure cases in the baseline KGE,
ontology-based negative sampling is effective for KGs consisting of 
real-world instances, such as
FB15k-237, while embedding-based negative sampling is powerful for
concept KGs such as WN18RR. 

{
\color{black}
\subsection{Performance as Training Progresses}

We plot the AUC-PR and MRR curve for training/validation, and testing in Fig. \ref{fig:aucpr} and 
Fig. \ref{fig:mrr}, respectively. We use AUC-PR to monitor the training of the classifiers. AUC-PR
starts to converge for both training and validation sets after 200 iterations. We record the 
link prediction results on the testing set every 100 iterations. Though the AUC-PR improves slightly
after 200 iterations, the MRR starts to converge after 600 iterations.
}

\end{document}